# Advocating Character Error Rate for Multilingual ASR Evaluation

**Thennal D K**
IIIT Kottayam
thennal21bcs14@iiitkottayam.ac.in

**Jesin James**
University of Auckland
jesin.james@auckland.ac.nz

**Deepa P Gopinath**
College of Engineering Trivandrum
deepapgopinath@cet.ac.in

**Muhammed Ashraf K**
University of Kerala
muhammed@keralauniversity.ac.in

## Abstract

Automatic speech recognition (ASR) systems have traditionally been evaluated using English datasets, with the word error rate (WER) serving as the predominant metric. WER's simplicity and ease of interpretation have contributed to its widespread adoption, particularly for English. However, as ASR systems expand to multilingual contexts, WER fails in various ways, particularly with morphologically complex languages or those without clear word boundaries. Our work documents the limitations of WER as an evaluation metric and advocates for the character error rate (CER) as the primary metric in multilingual ASR evaluation. We show that CER avoids many of the challenges WER faces and exhibits greater consistency across writing systems. We support our proposition by conducting human evaluations of ASR transcriptions in three languages—Malayalam, English, and Arabic—which exhibit distinct morphological characteristics. We show that CER correlates more closely with human judgments than WER, even for English. To facilitate further research, we release our human evaluation dataset for future benchmarking of ASR metrics. Our findings suggest that CER should be prioritized, or at least supplemented, in multilingual ASR evaluations to account for the varying linguistic characteristics of different languages.

## 1 Introduction

The majority of significant research on Automatic Speech Recognition (ASR) has been historically evaluated on a small set of English datasets (Jimerson et al., 2023; Baevski et al., 2020). In this context, while many evaluation metrics for ASR exist, the word error rate (WER) has garnered widespread use and has been adopted as the de facto metric for general ASR evaluation (Morris et al., 2004; Kim et al., 2021). WER as a metric enjoys many benefits: it is simple, easy to calculate, and has a direct interpretation as the probability of incorrect word recognition, or the expected proportion of incorrect words (Morris et al., 2004). As such, it serves as an adequate and easily reproducible evaluation metric for English ASR.

In recent years, with the increasing availability of training data and computational resources, industry research has shifted to the creation of large-scale multilingual ASR systems, able to transcribe hundreds of languages (Radford et al., 2023; Conneau et al., 2021). WER continues to be used as the de facto evaluation metric in this multilingual setting. However, the morphological characteristics of languages exhibit significant diversity, and the concept of a word error rate can fail in various ways.

The most straightforward examples are languages that have no word boundaries or separators, such as Thai and Chinese (Besacier et al., 2014). For such languages, the WER can only be applied after a subjective and potentially inaccurate word segmentation process (Wong et al., 2022). Modern research thus uses an alternative evaluation metric for these languages: the character error rate (CER).

CER, while not as ubiquitous as WER, enjoys widespread familiarity and frequent use in ASR evaluation (Radford et al., 2023; Conneau et al., 2021). Similar to WER, the CER is easy to calculate and to understand, and has a direct interpretation as the probability of incorrect character recognition. For a language like English, analytic and with clear word boundaries, a word-based metric can be argued to be more salient and useful than a character-based one. Further, a character-based metric is sensitive to the importance of a character in the writing system of a language. Despite these flaws, CER can be used as a consistent metric across languages, resistant to issues with morphological and orthographic variation, as we will show later in this work.

In this study, we present the assertion that modern ASR research should favor CER as the go-to metric for ASR evaluation over, or in addition to,

WER. We validate this assertion by documenting the ways in which WER fails as an evaluation metric, both theoretically and practically, while showing that CER is resistant to the same pitfalls. Further, we conduct subjective human evaluations for English, Malayalam, and Arabic, three languages from different families with varying morphological features. We correlate the human evaluations with the evaluations of each metric, providing further evidence. We also publicly release the human evaluations as a corpus, for future use in evaluating alternate ASR metrics, at `https://github.com/thennal10/asr-metric-evaluation`.

## 2 ASR Evaluation in a Morphologically Diverse Setting

In morphologically complex languages, orthographic changes at morpheme and affix boundaries are common, which may or may not be well-defined acoustically, especially in conversational speech. Beyond these morphological complexities, languages such as Thai and Vietnamese lack obvious separators between orthographic words (Besacier et al., 2014).

Table 1 provides examples of how WER might fail in the evaluation of morphologically complex languages. For each language, we provide two alternative orthographic hypotheses of the same sentences. Both sentences are grammatically valid and naturally occurring as verified by native speakers, and represent the same spoken sentence. For English, we note that they are in different dialects, and for Mandarin, one is written in traditional Chinese while the other is simplified Chinese. For all other languages, no large dialectical or orthographic difference exists between the two sentences.

An ideal ASR evaluation metric would treat both sentences equivalently, and the error between them would be zero. To compare, we take Hypothesis A as ground truth and calculate both the WER and CER as evaluation metrics. We note that in all cases, the WER is extremely high. The CER, while not zero, is always comparatively much lower, usually below 10% for most of the languages.

While these sentences were chosen to highlight the discrepancy, we stress again that these are naturally occurring sentences in the respective languages, and that the languages presented are a small and illustrative subset of all languages in which similar issues may arise.

## 3 Potential Solutions

### 3.1 Normalization

Normalizing before computing error rates is a standard practice to prevent unfair penalties on evaluation metrics, and have been found to enhance correlation with human judgement (Radford et al., 2023; Ali et al., 2019; Manohar and Pillai, 2024). However, this approach is not universally applicable, posing particular challenges for low-resource languages. Different scripts and languages demand specific normalization strategies, ranging from Unicode character normalization to alternate spelling equalization. Normalization schemes are largely unique and language-specific, and when working with hundreds of languages, such a normalization procedure becomes prohibitively difficult. Due to these challenges, authors working in multilingual settings often opt to either avoid normalization entirely or pursue simplified strategies for languages lacking established schemes, resulting in skewed and erroneous evaluations (Manohar and Pillai, 2024).

### 3.2 Alternative Metrics

The inadequacy of Word Error Rate (WER) as an evaluation metric has prompted exploration into alternative metrics, including letter or character error rate (Kurimo et al., 2006), phone error rate (PER), syllable error rate (SER, Huang et al., 2000), or morpheme error rate (MER, Ablimit et al., 2010). Additionally, inflectional word error rate (IWER, Bhanuprasad and Svenson, 2008; Karpov et al., 2011), and weighted word error rate (WWER, Nanjo and Kawahara, 2005) have been explored.

In natural language understanding, the focus on semantic correctness has led to the development of metrics based on semantic embeddings such as Semantic Distance (Kim et al., 2021) and BERTScore (Zhang et al., 2020), which exhibit strong correlations with downstream NLP task performance. Additionally, semantic metrics align most closely with human evaluations in the case of French ASR (Bañeras-Roux et al., 2023).

However, to ensure general adoption, the ideal ASR metric should reflect human judgment, while being simple to apply and dependant on language or application (Morris et al., 2004). Despite their applicability in the studied scenarios, the aforementioned metrics do not meet the criteria required for widespread use. Metrics like PER, SER, MER, and IWER, require language-specific processing schemes that might not be readily available for low-resource languages. MER and IWER are further only useful for information retrieval and connected speech recognition. Semantic metrics are highly complex and difficult to apply in a multilingual setting, requiring expensive and time-consuming training of embedding models and potential difficulties in achieving requisite performance levels for languages with limited training data. Thus, as a general multilingual evaluation metric for ASR, CER remains the only other option that fulfills the required criteria.

Table 1: Example sentences from different languages where CER and WER differ considerably

| Language | Hypothesis A | Hypothesis B | WER (%) | CER (%) |
|---|---|---|---|---|
| English | The colour drained from his face; he immediately apologised. | The color drained from his face. He immediately apologized. | 44.4 | 6.7 |
| Mandarin | 我认识很多中国人。 | 我認識很多中國人。 | 100.0 | 37.5 |
| Māori | He rorohiko utu nui tāku. | He roro hiko utu nui tāku. | 40.0 | 4.2 |
| Japanese | 妻に内緒で50万円の腕時計を買ってしまった。 | つまに内緒で50万円の腕時計を買ってしまった。 | 100.0 | 9.5 |
| Malayalam | അഞ്ചു ശതമാനം കൊടുക്കാമെന്നായിരുന്നു വാഗ്ദാനം. | അഞ്ച് ശതമാനം കൊടുക്കാം എന്നായിരുന്നു വാഗ്ദാനം. | 75.0 | 9.1 |
| Kannada | ಅವರು ಎಲ್ಲಿ ಇದ್ದಾರೆ. | ಅವರು ಎಲ್ಲಿದ್ದಾರೆ. | 66.7 | 11.1 |
| Arabic | ضَرَبَ زَيدٌ عَمْرًا. | ضَرَبَ زَيْدُنْ عَمْرَنْ. | 100.0 | 40.0 |
| Romanian | Maria nu îl cunoaște pe tatăl soțului ei. | Maria nuîl cunoaște pe tatăl soțuluiei. | 50.0 | 4.8 |
| Manipuri | ꯑꯃꯣꯅ ꯑꯣꯏꯕ ꯍꯦꯟꯅꯕꯗ ꯍꯧꯔꯦ꯫ | ꯑꯃꯣꯅ-ꯑꯣꯏꯕ ꯍꯦꯟꯅꯕꯗ ꯍꯧꯔꯦ꯫ | 50.0 | 5.0 |

## 4 Human Evaluation

To validate our assertion on the suitability of CER as a general ASR metric, we conducted a subjective evaluation with English, Malayalam, and Arabic ASR outputs. These languages were chosen to maximize morphological diversity. By calculating the correlation of human evaluation with WER and CER as automated evaluation metrics, we can quantify how well they align with human judgment.

Four widely used pre-trained ASR models—Whisper (Radford et al., 2023), XLSR-53 (Conneau et al., 2021, a multilingual version of Wav2Vec2 2.0, Baevski et al., 2020), Massively Multilingual Speech (MMS, Pratap et al., 2024), and SeamlessM4T (Barrault et al., 2023)—were selected for transcription generation to best capture practical transcription errors. Finetuned versions of XLSR-53 and Whisper were employed for Malayalam and Arabic due to performance considerations, while the original versions of MMS and SeamlessM4T sufficed for the evaluated languages without further finetuning.

For the subjective evaluation, we chose speech and corresponding transcriptions from the Common Voice 16.0 dataset (Ardila et al., 2020). The transcription from the dataset is taken as the ground truth. We selected 50 random speech-transcription pairs from the validation set of each language, while ensuring that the speech duration was at least five seconds to exclude very short sentences. Additionally, native speakers manually verified the selected sentences for naturalness and grammatical correctness.

Each sentence was transcribed by the four selected ASR models, resulting in four distinct transcriptions. These transcriptions, along with the original speech and ground truth, were organized into a survey. An example question is provided in Appendix E. The evaluators were instructed to rate the transcriptions, with the aid of the ground truth and original speech, on a continuous scale of 1 to 5. They were told to use small increments ($\leq 0.1$) for minor differences and large increments ($\geq 1$) for major differences that affect legibility and semantics. The evaluators were also instructed to have a mental ranking of the transcriptions and to score based on that ranking. The full text of the instructions is provided in Appendix D, and further details on the survey and the evaluators can be found in Appendix A.

| Language | Rating Correlation (\|%\|) | | Ranking Correlation (\|%\|) | | T-test |
|---|---|---|---|---|---|
| | WER | CER | WER | CER | $p$-value |
| English | 53.07 | **54.78** | 69.98 | **74.91** | $1.60 \times 10^{-12}$ |
| Malayalam | 34.91 | **41.54** | 47.32 | **51.15** | $5.19 \times 10^{-3}$ |
| Arabic | 32.59 | **32.86** | 40.93 | **46.42** | $1.26 \times 10^{-13}$ |

Table 2: Calculated correlation metrics (displayed as an absolute percentage) from the human evaluation. Rating correlation refers to the direct correlation between scores and metrics while ranking correlation is between rankings produced by scores and metrics.

## 5 Results & Discussion

The survey for each language was completed by 20 evaluators proficient in the respective language. To quantify the effectiveness of WER and CER, we associate the metric with the human evaluation. We acknowledge that the raw ratings are highly subjective, with different evaluators having a different distribution of scores. However, as the evaluators were instructed to score with a mental ranking in mind, we generated rankings for each set of 4 transcriptions. These rankings reduce the variation between evaluators, providing a more reliable measure.

Specifically, we calculate a correlation measure between a) the raw metric output and the human ratings, and b) the ranking generated from the metric and the ranking generated from the human ratings. To evaluate the correlation between the raw metric and rating, the Pearson correlation coefficient was taken. For rank correlation, we took each set of four alternative transcriptions, ranked them according to the human ratings and the metric, and took Spearman's rank correlation coefficient between the two rankings. We further took the mean of these coefficients to produce a single rank correlation measure.

The results are aggregated in Table 2. For both the rating and rank correlation, CER scores higher than WER in all three languages. In the case of rank correlation, the difference between CER and WER is consistent and significant, averaging $+4.75\%$ in favor of CER. Rankings remove much of the subjectivity and variance from the evaluation and so the rank correlation results should be considered with more weight.

Further, taking the rank correlation coefficient of each question as a random variable, we conduct a one-sided repeated Student's T-test to calculate the probability that the CER correlation is higher than the WER correlation by chance. The associated $p$-values are reported in the table. As they are all much less than the generally accepted threshold of $0.05$, we can dismiss the null hypothesis that this discrepancy in correlation could be caused by random chance.

Normalization measures may also affect the correlation. While no standard normalization scheme for Arabic and Malayalam exist, many exist for English and are usually deployed before WER evaluation. To emulate real-world use cases more closely, we also deploy the normalization scheme detailed by Radford et al. (2023) for English and recalculate the correlation. We find consistent results with regards to our hypothesis: a ranking correlation of $35.92$ and $51.39$ for WER and CER respectively ($31.82$ and $43.16$ in case of rating correlation).

We also calculate the interrater reliability by calculating Kendall's W between all raters for each question and averaging the result. A $0$ denotes no agreement, while a $1$ means complete agreement. For English and Malayalam, the reliability is $0.6466$ and $0.5598$, indicating general agreement in ranking. For Arabic, Kendall's W is a lower $0.3438$, but still indicates that there was reasonable agreement.

We posit that the results agree with our hypothesis: that across languages of different families, with different morphological features, WER is a consistently subpar metric compared to CER. Even in the case of English, CER correlates better with human evaluators based on both raw scores and rankings. A similar study done on French ASR corroborates our findings (Bañeras-Roux et al., 2023).

## 6 Conclusions

An ideal ASR metric should reflect human judgment while being simple to apply and not dependent on language or application (Morris et al., 2004). Given the results of the human evaluation we conducted and our analysis of the language-based issues WER faces, we assert that CER qualifies as a better multilingual ASR metric. Further, the preexisting popularity of CER puts it in an advantageous position for wide adoption.

## 7 Limitations

Although they are from different families and morphological types, our human evaluation considers only three languages. We also only consider evaluating two metrics, WER and CER. However, we justify our choice in Section 3.2, and leave a comparison with other metrics for future work. As we do not specify any particular end use-case in our instructions to human evaluators, these results may not be valid for certain downstream tasks which prioritize different aspects of the transcripts. Nonetheless, we posit that the consistency and magnitude of CER's correlation with human judgement in comparison to WER makes a strong general case.

## A Human Evaluation Survey

The evaluators were voluntary participants, recruited by advertising the survey at large public universities in Kerala and New Zealand. Interested individuals could access a provided link, which directed them to the Participant Information Sheet (PIS). The PIS clearly outlines the nature of the study and the use of the data provided, and the full text is provided in Appendix B. If they chose to proceed with the survey after reviewing the PIS, the first page of the survey provides a consent form (full text given in Appendix C), which they are required to agree to in order to continue with the rest of the survey.

The evaluators were asked to rate their own proficiency with the language in question. For Malayalam and English, all evaluators were native speakers. In the case of Arabic, due to the unavailability of native speakers in the region where the test was conducted, we included advanced learners of the language. All evaluators, regardless of the survey, were fluent in English.

## B Participant Information Sheet

**Project Title:** Perception test for Transcriptions in English, Malayalam, and Arabic
**Principal Investigators:** Thennal D K, Jesin James, Deepa P Gopinath, Muhammed Ashraf K

Hello,
You are receiving this Participant Information Sheet because you have expressed interest in participating in this experiment. The goal of this study is to gather human evaluations of transcription quality across various languages, which will help us understand how individuals perceive transcription accuracy.

**Survey Structure:**
The survey will present you with speech samples and their corresponding transcriptions in English, Malayalam, or Arabic depending on the language you have chosen to participate in. You will be asked to rate the quality of the transcriptions on a scale of 1 to 5, considering factors such as legibility, grammar, punctuation, and overall correctness.

The transcriptions you will see have been generated from real speech data. Your task is to evaluate how well the transcription matches the speech, based on the provided text and audio. No prior knowledge of transcription systems is required, and you will not be informed of the source of the transcriptions.

This survey is web-based and can be completed at your convenience using a computer and headphones or earphones. It will consist of:

- **Demographics Questionnaire:** We will ask some general questions, such as your age and familiarity with the language the survey is based on.
- **Transcription Evaluation:** You will listen to speech recordings and rate the accompanying transcriptions in terms of their overall accuracy and quality.

The survey is designed to take approximately 30–45 minutes, although this may vary depending on the individual. You can take as much time as needed.

**Confidentiality and Data Handling:**
Participation in this survey is entirely voluntary. You may withdraw at any time by closing the survey page. No identifying personal information will be collected. All responses will remain anonymous, and the data collected will be stored securely in University of Auckland research storage systems, accessible only to the research team.

If you would like to receive the results of the study, you can provide your email address at the end of the survey. This will be stored separately from your responses to ensure your anonymity. Your email address will not be shared with third parties.

By participating and submitting the survey, you acknowledge that:

- You are at least 18 years old.
- You have read and understood the information supplied here.
- You consent to participate under the terms outlined above.

If you have any questions or concerns about this research, please feel free to contact the research team at `[Contact details]`.

Thank you for your participation.

## C Consent Form ( Survey Participants)

This form will be held for a period of 6 years.
**Project Title:** Perception test for transcriptions in English, Malayalam, and Arabic
**Principal Investigators:** Thennal D K, Jesin James, Deepa P Gopinath, Muhammed Ashraf K

By participating in this survey, I acknowledge the following:

1. I have read the Participant Information Sheet and understand the nature of the research and my involvement.
2. I have had the opportunity to ask questions and have received satisfactory answers.
3. My participation in the study is voluntary.
4. I am at least 18 years old.
5. I understand that I will be asked to evaluate transcriptions based on speech samples and provide ratings, as described in the Participant Information Sheet.
6. I understand that my participation is anonymous, and no identifying personal information will be collected.
7. I understand that I may withdraw from the study at any time by closing the survey page, and my responses will not be saved.
8. I understand that once the survey is submitted, I will not be able to withdraw my responses.
9. I understand that I will not be able to review or edit my responses after submission.
10. I understand that my data will be stored securely and used for academic research purposes.

Please select one of the following options:

- I consent, begin the survey.
- I do not consent, I do not wish to participate.

## D Evaluator Instructions

There will be 50 questions. Each question will provide a speech clip, and an accurate transcript of that speech, referred to as the **ground truth**. Four different alternative transcriptions are also provided which may be erroneous. Your task is to rate each of the four alternative transcriptions on how accurate they are. The comparison should be made with the speech clip, and the ground truth is provided as a reference and example of a potentially accurate transcription.

The rating scale is continuous between 1 and 5. The criteria for each integer rating are as follows:

| Rating | Associated Meaning |
|---|---|
| 1 | Transcript is completely unrelated to the speech clip. Gibberish, nonsensical transcripts are reserved for this rating. |
| 2 | Transcript is filled with major mistakes. The original sentence cannot be parsed, and only a few words are correctly transcribed. |
| 3 | Transcript does not accurately transcribe many specific words and contains mistakes, but you can still parse the original sentence with some difficulty. |
| 4 | Transcript conveys the meaning of the spoken sentence and transcribes most of it accurately, with some minor mistakes. |
| 5 | Transcript is completely accurate to the speech clip, both grammatically and semantically. |

Table 3: Rating Scale

As the rating scale is continuous, you can rate between these integer ratings, such as 4.35, 1.24, etc. You may use your intuition to rate the transcripts. As a rough guideline, a 0.1 difference in rating indicates a minor difference that does not particularly affect the meaning of the sentence (e.g., punctuation). A 0.01 difference is only used to indicate a stylistic preference: both transcripts should be equally accurate. If two transcripts are perfectly equivalent, character for character, you can give them the same score. Otherwise, **all transcripts must be given a different score**. This is to force a ranking between the transcriptions.

The following provides examples of potential questions and answers:

### Q1

**Ground Truth:** He went into the time machine and never came back.

| No. | Transcript | Rating |
|---|---|---|
| 1 | He went into the time machine and never came back. | 4.90 |
| 2 | He went into the time-machine and never came back. | 4.99 |
| 3 | He went into the time machine and never came back. | 5.00 |
| 4 | E went into time machine and never game back. | 3.23 |

Table 4: Example Question 1

Here, the evaluator looks at the ground truth and determines that "time machine" without the hyphen is the preferred way to write it, and so gives an edge to transcript number 3 with a full rating of 5, as it is a perfect match. Since the only difference in transcript 2 is the hyphen, a stylistic choice, the evaluator gives it 4.99. Transcript 1 is missing punctuation, so the evaluator docks 0.1 points, giving it 4.90. Transcript 4 has significant issues, but one can still guess the underlying sentence, so the evaluator rates it a bit above 3 according to their intuition.

### Q2

**Ground Truth:** Is that so?

| No. | Transcript | Rating |
|---|---|---|
| 1 | Is that so! | 5.00 |
| 2 | I have no? | 1.64 |
| 3 | In that goal | 1.45 |
| 4 | Is that | 2.87 |

Table 5: Example Question 2

Here, the only difference between transcript 1 and the ground truth is punctuation. However, the evaluator hears the speech clip and decides that an exclamatory tone is also a completely valid reading of the spoken sentence. As such, (and without any competition from the other transcripts), the evaluator decides to rate it the full 5 points. The other three transcripts are all extremely erroneous. The third sentence gets at least two of the words, but the original meaning of the sentence cannot be parsed, so the evaluator gives it a score a bit below 3. For the other two, the evaluator decides that while they are both bad transcriptions, "I have no?" is better than "In that goal" and decides to give them scores around 1.5, with a 0.2 difference reflecting the difference.

## E Example Question

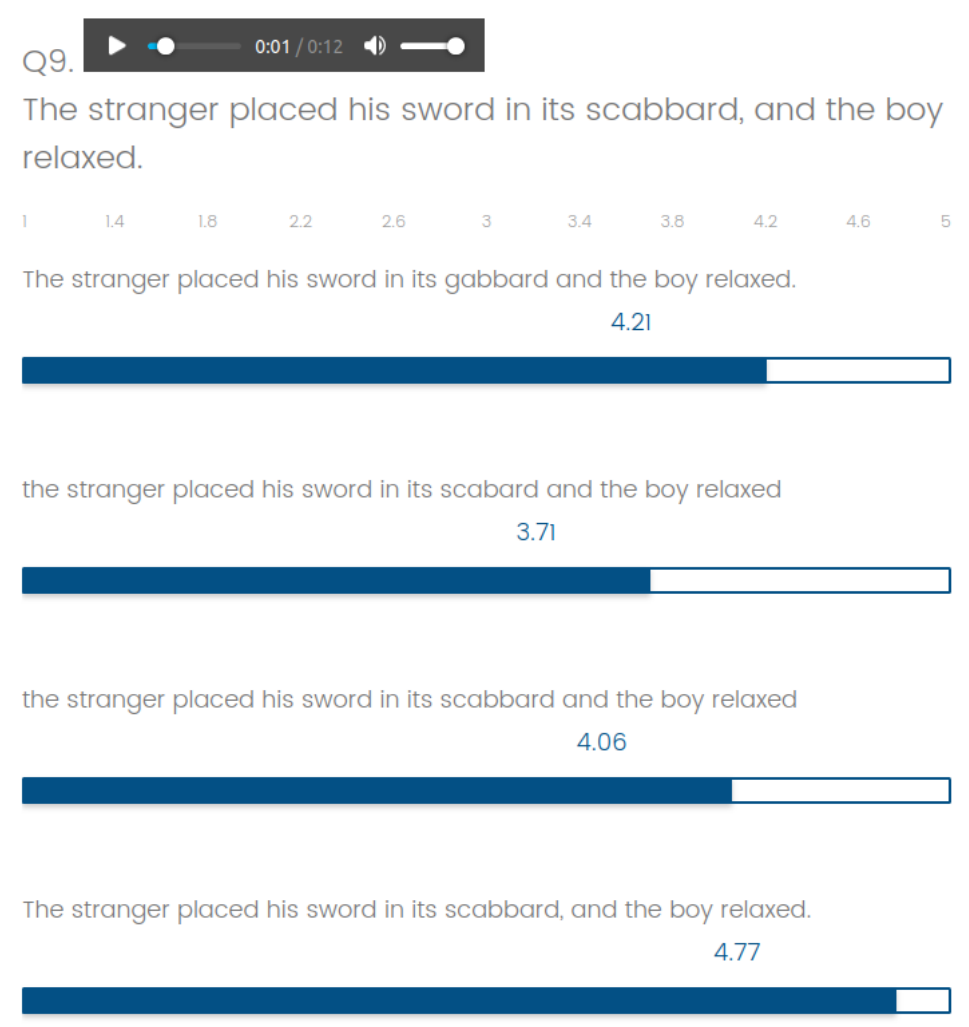

Figure 1: Example question, filled out, showing an audio-based transcription evaluation task.

Fig. 1 shows a sample question from the evaluation survey. We also transcribe the question below.

### Q9

`<audio player>`

The stranger placed his sword in its scabbard, and the boy relaxed.

| Transcript | Rating |
|---|---|
| The stranger placed his sword in its gabbard and the boy relaxed. | 4.21 |
| the stranger placed his sword in its scabard and the boy relaxed | 3.71 |
| the stranger placed his sword in its scabbard and the boy relaxed | 4.06 |
| The stranger placed his sword in its scabbard, and the boy relaxed. | 4.77 |

Table 6: Transcripts and their corresponding ratings for Question 9.